\documentclass{article}

\usepackage{amsmath}
\usepackage{microtype}
\usepackage{graphicx}
\usepackage{subfigure}
\usepackage{booktabs} % for professional tables
\usepackage{comment}

\usepackage{hyperref}

\usepackage[accepted]{icml2024}

\usepackage{amsmath}
\usepackage{amssymb}
\usepackage{mathtools}
\usepackage{amsthm}

\usepackage[capitalize,noabbrev]{cleveref}

\theoremstyle{plain}

\theoremstyle{definition}

\theoremstyle{remark}

\usepackage[textsize=tiny]{todonotes}

\icmltitlerunning{Transferability of deep learning models for climate downscaling}

\begin{document}

\twocolumn[
\icmltitle{Evaluating the transferability potential of deep learning models for climate downscaling}




\begin{icmlauthorlist}
\icmlauthor{Ayush Prasad}{comp,hel}
\icmlauthor{Paula Harder}{comp}
\icmlauthor{Qidong Yang}{mit}
\icmlauthor{Prasanna Sattegeri}{ibm}
\icmlauthor{Daniela Szwarcman}{ibm}
\icmlauthor{Campbell Watson}{ibm}
\icmlauthor{David Rolnick}{comp}
\end{icmlauthorlist}

\icmlaffiliation{hel}{University of Helsinki, Helsinki, Finland}
\icmlaffiliation{comp}{Mila Quebec AI Institute, Montreal, Canada}
\icmlaffiliation{ibm}{IBM Research, US/Brazil}
\icmlaffiliation{mit}{Massachusetts Institute of Technology, Cambride, MA, US}

\icmlcorrespondingauthor{Ayush Prasad}{ayush.prasad@mila.quebec}

\icmlkeywords{Machine Learning, ICML}

\vskip 0.3in
]



\printAffiliationsAndNotice{} % otherwise use the standard text.

\begin{abstract}
Climate downscaling, the process of generating high-resolution climate data from low-resolution simulations, is essential for understanding and adapting to climate change at regional and local scales. Deep learning approaches have proven useful in tackling this problem. However, existing studies usually focus on training models for one specific task, location and variable, which are therefore limited in their generalizability and transferability. In this paper, we evaluate the efficacy of training deep learning downscaling models on multiple diverse climate datasets to learn more robust and transferable representations. We evaluate the effectiveness of architectures zero-shot transferability using CNNs, Fourier Neural Operators (FNOs), and vision Transformers (ViTs). We assess the spatial, variable, and product transferability of downscaling models experimentally, to understand the generalizability of these different architecture types. %We also explore the use of fine-tuning on the pre-trained models to further improve performance. Our results demonstrate that the proposed pre-training approach improves climate downscaling performance in most scenarios.
\end{abstract}

\section{Introduction}
Climate downscaling refers to the creation of synthetic high-resolution climate data from coarse-resolution data. Downscaling can be useful when raw datasets, either reanalysis or climate model outputs, have a spatial resolution that is insufficient for applications requiring climate information at local scales. Running climate models at high resolutions is computationally intensive and often infeasible due to the enormous computational resources required. Downscaling addresses this challenge by providing high-resolution information with the need for time-intensive simulations. Deep learning is increasingly being used in climate downscaling, including methods such as Convolutional Neural Networks (CNNs) \cite{rampal2022high, harder2023hard}, Fourier Neural Operators (FNOs) \cite{yang2023fourier}, and Conditional Normalizing Flows \cite{winkler2023towards}. These models have demonstrated the ability to capture complex spatial and temporal patterns in climate data, leading to improved downscaling accuracy compared to traditional statistical methods.

However, existing studies primarily focus on training deep learning models on a single dataset, which can result in overly specialised models that fail to capture the diverse characteristics of climate data across different sources and resolutions, limiting their generalizability and transferability to different climate variables, spatial regions, and temporal periods. %Consequently, there is a need to explore approaches that can enhance the robustness and adaptability of deep-learning models for climate downscaling.
In this work, we develop and evaluate approaches to address this challenge by training deep learning models on multiple climate datasets and evaluating them on different tasks (zero-shot transfer). We hypothesize that exposing the models to a diverse range of climate information during training will enable them to learn more comprehensive and transferable representations of climate patterns by capturing the underlying structures and relationships present across various datasets.%, models are expected to exhibit improved generalization capabilities and adaptability to new climate variables, spatial regions, and temporal periods.

In this work, we consider the transferability of several deep learning architectures, including CNNs, Fourier Neural Operators (FNOs), and Transformers. We train these models on a combination of climate reanalysis data products, which include a set of climate variables at different spatial resolutions and temporal frequencies. By leveraging the complementary information present in these datasets, we aim to enhance the models' ability to learn robust and transferable representations of climate patterns.

We conduct a series of experiments that assess the downscaling models' spatial, variable, and product transferability. Furthermore, we explore the impact of fine-tuning the pre-trained models on the target dataset for the case where zero-shot performance is not sufficient. %Fine-tuning allows the models to adapt to the specific characteristics of the target dataset while leveraging the knowledge gained during pre-training. 

% The main contributions of this paper are as follows:

% \begin{enumerate}
% \item We investigate the zero-shot transferability of deep learning methods for climate downscaling.

% \item We conduct experiments to assess the spatial, variable and product transferability.

% \item We compare different models such as CNNs, Fourier Neural Operators, and Transformers based on their generalization performance in the context of two different downscaling factors.

% \end{enumerate}

\section{Data}
\label{sec:data}
In this study, we utilize multiple climate datasets to pre-train our deep learning models for climate downscaling. The datasets are selected based on their spatial resolution, temporal frequency, and the variety of climate variables they provide. By leveraging diverse datasets, we aim to capture a wide range of climate patterns and enhance the generalizability of the models. The following datasets are used in our experiments:

\subsection{ERA5}
ERA5 \cite{hersbach2020era5} is a state-of-the-art atmospheric reanalysis dataset produced by the European Centre for Medium-Range Weather Forecasts (ECMWF) \cite{hersbach2020era5}. It provides hourly data at a spatial resolution of 0.25$^{\circ}$ $\times$ 0.25$^{\circ}$ on a single pressure level. We consider the following climate variables from ERA5:
2-meter temperature,  2-meter dewpoint temperature, u-10 component of wind, v-10 component of wind, mean sea level pressure, and total precipitation. The total number of samples used from the ERA5 dataset is 50,000.

\subsection{MERRA2}
MERRA2 (Modern-Era Retrospective Analysis for Research and Applications, Version 2) \cite{gelaro2017modern} is an atmospheric reanalysis dataset. It provides data at a spatial resolution of 0.5$^{\circ}$ $\times$ 0.625$^{\circ}$. We utilize the following climate variables from the MERRA-2 M2T1NXFLX, Surface Flux Diagnostics V5.12.4 dataset:
Total precipitation, surface wind speed, surface air temperature, surface eastward wind, surface northward wind. The total number of samples used from the ERA5 dataset is 50,000.

\subsection{NOAA CFSR}
The NOAA Climate Forecast System Reanalysis (CFSR) is a global coupled atmosphere-ocean-land surface-sea ice system dataset \cite{saha2010ncep}. It provides data at a spatial resolution of 0.5$^{\circ}$ $\times$ 0.5$^{\circ}$. We consider the following climate variables from the CFSR dataset:
Precipitable water at the entire atmosphere layer, downward longwave radiation flux, surface temperature, and pressure at mean sea level. Our dataset comprises of 30,000 samples.

\subsection{NorESM}
Our NorESM data set is based on the second version of the Norwegian Earth System Model (NorESM2) \cite{bentsen2013norwegian}, which is a coupled Earth System Model developed by the NorESM Climate modeling Consortium (NCC), based on the Community Earth System Model, CESM2. We build our data set on two different runs: NorESM-MM which has a 1-degree resolution and NorESM2-LM which has a 2-degree resolution for atmosphere and land components. We use the temperature at the surface (tas) and a time period from 2015 to 2100. The data is cropped into patches of $64 \times 64$ (HR) and $32\times 32$ pixels (LR). The scenarios ssp126 and ssp585 are used for training and ssp245 for testing with a total sample size of 37152.

\subsection{Data Preprocessing}
To ensure consistent input to the deep learning models, we scale all variables to have zero mean and unit variance. This normalization step helps in improving the convergence and stability of the training process.

Following a common approach \citep{stengel_2020,spatio_temp2, spatio_temp1}, we create low-resolution (LR) and high-resolution (HR) pairs for training and evaluation by applying average pooling to the high-resolution data. Specifically, we use a pooling kernel size of 2x2 to downsample the HR data to generate the corresponding LR data for the 2x scale factor, and a pooling kernel size of 8x8 for the 8x scale factor.

\section{Methods}
\label{sec:methods}
\subsection{Convolutional Neural Networks}
Convolutional Neural Networks (CNNs) have performed well in image super-resolution tasks, including climate downscaling \cite{dong2015image, yang2019deep}. The key components of CNNs are convolutional layers, which apply learnable filters to extract local features from the input data. In our CNN model, we adopt a residual architecture inspired by the Super-Resolution Convolutional Neural Network (SRCNN) \cite{dong2015image}. The model consists of multiple residual blocks, each containing two convolutional layers with a kernel size of 3×3 and 64 filters, followed by a Rectified Linear Unit (ReLU) activation function. The number of residual blocks is set to 16. The output of the last residual block is passed through a final convolutional layer to generate the high-resolution climate variable.

\subsection{Fourier Neural Operators}
Fourier Neural Operators (FNOs) are a recently proposed class of models that learn mappings between function spaces by operating in the Fourier domain \cite{li2020fourier}. The key idea behind FNOs is to represent the input and output functions using their Fourier coefficients and learn the mapping between the coefficients using a neural network. In our FNO model, we use the architecture developed by \cite{yang2023fourier} for downscaling climate data. The input low-resolution climate variable is first transformed into the Fourier domain using the Fast Fourier Transform (FFT). The Fourier coefficients are then processed by a series of Fourier layers, each consisting of a linear transformation followed by a non-linear activation function. The linear transformation is performed in the Fourier domain, which allows for efficient and resolution-invariant learning. The output of the last Fourier layer is transformed back to the spatial domain using the Inverse Fast Fourier Transform (IFFT) to obtain the high-resolution climate variable. The number of Fourier layers is set to 4, and the number of modes (i.e., the number of Fourier coefficients) is set to 12 in each dimension.

\subsection{CNN-ViT Hybrid Model}
Vision Transformers (ViTs) have recently gained popularity in computer vision tasks due to their ability to capture global context and model long-range dependencies \cite{dosovitskiy2020image}. However, ViTs often require large amounts of training data and may struggle with localized features. To address these limitations, we use a hybrid architecture that combines CNNs and ViTs, following recent works on integrating CNNs and ViTs \cite{fang2022hybrid}. The CNN-ViT model first uses a convolutional stem to extract local features from the input low-resolution climate variable. The convolutional stem consists of two convolutional layers with a kernel size of 3$\times$3 and 64 filters, followed by a ReLU activation function. The output of the convolutional stem is then reshaped and passed through a ViT module to capture global interactions. The ViT module follows the standard Transformer architecture \cite{vaswani2017attention}, consisting of multiple multi-head self-attention layers and feed-forward networks. The number of Transformer layers is set to 4, with 4 attention heads and a hidden dimension of 256. The output of the ViT module is reshaped and passed through a final convolutional layer to generate the high-resolution climate variable.

\subsection{Training}
All models are trained using the Adam optimizer with a learning rate of 0.001 and a batch size of 32. We train the models for 150 epochs, which each individually takes approximately 7 hours on a single Nvidia V100 GPU.

\section{Experiments and Results}
\label{sec:experiments}
\subsection{Spatial Transferability}
In this experiment (Table \ref{spatial}), we train the models on the ERA5, MERRA2 and the NOAA CFSR dataset in the DACH region (Germany, Austria, and Switzerland) between 2010 and 2023. The DACH region is defined by the coordinates 45$^\circ$N to 55$^\circ$N latitude and 5$^\circ$E to 15$^\circ$E longitude. We then evaluate the trained models' performance on a region in North America, defined by the coordinates 35$^\circ$N to 50$^\circ$N latitude and 70$^\circ$W to 125$^\circ$W longitude, covering a portion of the continental United States. The models are evaluated on a subset of the North American data. The goal of this experiment is to assess the spatial transferability of the models, i.e., their ability to generalize and perform well on a geographic region distinct from the region they were trained on. The results in Table \ref{spatial}, demonstrate that all three deep learning models outperform the bicubic interpolation baseline in both the 2$\times$ and 8$\times$ downscaling scenarios. The CNN-ViT hybrid model achieves the highest $R^2$ scores and lowest MSE values among the compared models. The FNO model also had strong performance, closely following the CNN-ViT model. The CNN model, while still performing better the bicubic baseline, has lower performance compared to the other models.

\begin{table}[htb]
    \centering
    \caption{Comparison of spatial transferability performance for different models in terms of $R^2$ and MSE. The models were trained on the DACH region and evaluated on a subset of North America for 2$\times$ and 8$\times$ downscaling factors}
    \label{spatial}
    \vskip 0.15in
    \begin{small}
    \begin{sc}
    \begin{tabular}{l|cc|cc}
        \toprule
        & \multicolumn{2}{c}{2$\times$} & \multicolumn{2}{c}{8$\times$} \\ 
        Model & $R^2$ & MSE & $R^2$ & MSE \\ \midrule
        CNN & 0.94 & 0.005 & 0.88 & 0.012 \\
        FNO & 0.96 & 0.004 & 0.90 & 0.010 \\
        CNN-ViT & \textbf{0.97} & \textbf{0.003} & \textbf{0.92} & \textbf{0.008} \\
        Bicubic & 0.92 & 0.007 & 0.86 & 0.015 \\
        \bottomrule
    \end{tabular}
    \end{sc}
    \end{small}
    \vskip -0.1in
\end{table}

\subsection{Variable Transferability}
In this experiment (Table \ref{variable}), we first train the models on temperature, wind, and precipitation variables from all the climate reanalysis datasets (ERA5, MERRA2, NOAA CFSR). None of the flux variables, such as downward longwave radiation flux, were included in the training data. After training, we evaluate the models' performance on the downward longwave radiation flux variable from the NOAA CFSR dataset. This variable was intentionally excluded from the training data. The aim is to evaluate whether the models can effectively generalize and transfer the learned representations to predict an unseen climate variable distinctly different from the variables they were exposed to during training. The results in Table \ref{variable} show that the FNO model achieves the highest $R^2$ scores and lowest MSE values for both 2$\times$ and 8$\times$ downscaling. The CNN-ViT model performs slightly better than the CNN model for 2$\times$ downscaling, but the CNN model outperforms the CNN-ViT model for 8$\times$ dowmscaling. The FNO model demonstrates the best transferability to the unseen downward longwave radiation flux variable, which was not included in the training data.

\begin{table}[h]
\caption{Comparison of variable transferability performance for different models in terms of $R^2$ and MSE. The models were trained on temperature, wind, and precipitation variables and evaluated on the unseen downward longwave radiation flux variable for 2$\times$ and 8$\times$ downscaling factors}
\label{variable}
\vskip 0.15in
\begin{center}
\begin{small}
\begin{sc}
\begin{tabular}{l|cc|cc}
\toprule
& \multicolumn{2}{c}{2$\times$} & \multicolumn{2}{c}{8$\times$} \\ 
Model & $R^2$ & MSE & $R^2$ & MSE \\ \midrule
CNN & 0.93 & 0.004 & 0.87 & 0.013 \\
FNO & \textbf{0.95} & \textbf{0.004} & \textbf{0.89} & \textbf{0.011} \\
CNN-ViT & 0.94 & 0.005 & 0.86 & 0.014 \\
Bicubic & 0.92 & 0.007 & 0.85 & 0.016 \\
\bottomrule
\end{tabular}
\end{sc}
\end{small}
\end{center}
\vskip -0.1in
\end{table}

\begin{figure}
\vskip 0.15in
    \centering
    \includegraphics[width = 3.4in]{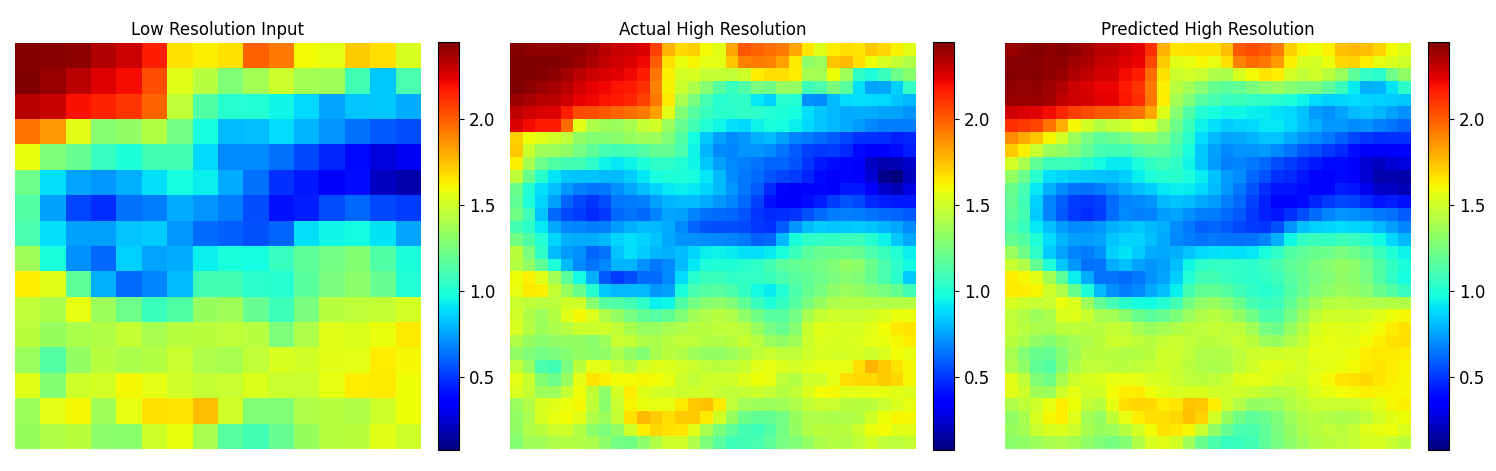}
    \caption{An example of 2x downscaling of ERA5 2m temperature using the CNN-ViT hybrid model. Here the model was trained on the DACH region and evaluated on a subset of North America.}
    \label{fig:enter-label}
\vskip -0.1in
\end{figure}

\subsection{Product Transferability}
In this experiment (Table \ref{product}), we train the models on ERA5 and MERRA2 datasets and evaluate their performance on the NOAA CFSR dataset without fine-tuning, i.e., their zero-shot performance. The goal is to examine the transferability of the models across different climate data products. The results in Table \ref{product} show that the CNN-ViT model achieves the highest R2 scores and lowest MSE values for both 2$\times$ and 8$\times$ downscaling. The FNO follows closely, outperforming the CNN.

\begin{table}[h]
\caption{Comparison of product transferability performance for different models in terms of $R^2$ and MSE. The models were trained on ERA5 and MERRA2 datasets and evaluated on the NOAA CFSR dataset for 2$\times$ and 8$\times$ downscaling}
\label{product}
\vskip 0.15in
\begin{center}
\begin{small}
\begin{sc}
\begin{tabular}{l|cc|cc}
\toprule
& \multicolumn{2}{c}{2$\times$} & \multicolumn{2}{c}{8$\times$} \\ 
Model & $R^2$ & MSE & $R^2$ & MSE \\ 
\toprule
CNN & 0.92 & 0.007 & 0.86 & 0.015 \\
FNO & 0.94 & 0.005 & 0.89 & 0.011 \\
CNN-ViT & \textbf{0.95} & \textbf{0.004} & \textbf{0.90} & \textbf{0.010} \\
Bicubic & 0.88 & 0.009 & 0.84 & 0.017 \\
\bottomrule
\end{tabular}
\end{sc}
\end{small}
\end{center}
\vskip -0.1in
\end{table}

\subsection{Two Simulation Transferability}
In this experiment (Table \ref{transfer}), we evaluate the models' performance on real-world low-resolution (LR) and high-resolution (HR) data pairs obtained from the Norwegian Earth System Model (NorESM) \cite{bentsen2013norwegian}. Unlike the previous experiments where LR data was generated by average pooling of the HR data, here we use the actual coarse-resolution data from NorESM as LR inputs, and the corresponding high-resolution data as HR targets. The LR data has a spatial resolution of 2.5$^\circ$ $\times$ 1.9$^\circ$, while the HR data has a resolution of 0.25$^\circ$ $\times$ 0.25$^\circ$.
Initially, we trained the models on all the reanalysis datasets and evaluate their performance on the NorESM data without any additional fine-tuning. However, the results were just slightly better than interpolation, indicating the need for domain adaptation to the specific characteristics of the NorESM data.
To improve performance, we perform fine-tuning on the pre-trained models on 30 percent of the NorESM training data. The results in Table \ref{transfer} show that when the models are directly applied to the NorESM dataset without fine-tuning (zero-shot), their performance is relatively poor compared to the previous experiments. The FNO and CNN-ViT models perform slightly better than the CNN model and the bicubic interpolation baseline in the zero-shot setting. After fine-tuning (FT) the pre-trained models on the NorESM data, their performance improves. The CNN-ViT model achieves the highest $R^2$ scores and lowest MSE values for both 2$\times$ and 8$\times$ downscaling factors, followed by the FNO model. The CNN architecture also shows improvement after fine-tuning.

\begin{table*}[h]
\caption{Comparison of real-world LR-HR (NorESM) transferability performance for different models in terms of $R^2$ and MSE. The models were initially pre-trained on all reanalysis datasets and then evaluated on the NorESM dataset without fine-tuning (zero-shot, ZS) and with fine-tuning (FT) on 30 percent of the NorESM data for 2$\times$ and 8$\times$ downscaling factors. }
\label{transfer}
\vskip 0.15in
\begin{center}
\begin{small}
\begin{sc}
\begin{tabular}{lccccc}
\toprule
Type  & Scale & CNN ($R^2$/MSE) & FNO ($R^2$/MSE) & CNN-ViT ($R^2$/MSE) & Bicubic ($R^2$/MSE) \\
\midrule
ZS & 2x & 0.91/0.0065 & \textbf{0.93/0.0055} & 0.93/0.0082 & 0.92/0.0060 \\
ZS & 8x & 0.87/0.0120 & 0.89/0.0110 & \textbf{0.91/0.0100} & 0.90/0.0095 \\
\midrule
FT & 2x & 0.97/0.0030 & 0.98/0.0025 & \textbf{0.99/0.0020} & 0.96/0.0040 \\
FT & 8x & 0.92/0.0080 & 0.94/0.0060 & \textbf{0.96/0.0050} & 0.93/0.0090 \\
\bottomrule
\end{tabular}
\end{sc}
\end{small}
\end{center}
\vskip -0.1in
\end{table*}

\section{Conclusion and Future Work}
\label{sec:conclusion}
In this paper, we investigate the performance of deep learning models for climate downscaling when evaluated on different data than orignally trained on. We assess the spatial, variable and product transferability of the different models. The results indicate that training on diverse datasets leads to models that have good zero-shot capability, performing better than simple bicubic interpolation on many tasks. 
All three model types, CNNs, FNOs and Transformers outperform the baseline on all tasks zero-shot, apart from the the two-simulation-NorESM setup, where additional fine-tuning is necessary. For spatial and product transferability the CNN-ViT shows the best performance across architectures, whereas for variable transferability the FNO-based model has the highest scores.

In the two-simulation transferability experiment, we observe relatively poor performance when directly transferring models without adaptation. However, further fine-tuning the pre-trained models on a subset of the target dataset significantly improved their accuracy. Overall, the results indicate that pre-training on multiple climate datasets, combined with fine-tuning, is an effective strategy for developing deep learning models for climate downscaling. Next steps of this work are the application of pre-training and fine-tuning for variable, spatial and product transferability, as it was done for the NorESM case. This could be evaluated against directly training on the target dataset. Here it could be valuable to see how the choice, number, and sizes of pre-training datasets affect the transferability performance as well of the number of data points chosen for fine-tuning.
%A successful training strategy including pre-training on large datasets can improve 
%fine-tuning like noresm for the other tasks as well
%pre-training+fine-tuning vs directly training on the target
%There are several promising directions for future work in this area. One natural next step would be to investigate the impact of fine-tuning on the other transferability tasks, similar to what was done for the NorESM experiment. This could help determine the effectiveness of fine-tuning in improving the transferability of deep learning models for climate downscaling. Another important direction is to compare the performance of models that are pre-trained on multiple datasets and then fine-tuned on the target dataset versus models that are directly trained on the target dataset from scratch. This comparison could guide the development of optimal training strategies for climate downscaling models. Furthermore, it would be valuable to understand the role of the number and sizes of datasets used for pre-training. 

\bibliography{example_paper}
\bibliographystyle{icml2024}
\end{document}